\documentclass[letterpaper]{article} 
\usepackage{aaai25} 
\usepackage{times} 
\usepackage{helvet}  
\usepackage{courier}  
\usepackage[hyphens]{url} 
\usepackage{graphicx}
\usepackage{natbib}  
\usepackage{caption}
\usepackage{algorithm}
\usepackage{algorithmic}

\usepackage{multirow}
\usepackage{colortbl}

\usepackage{float}

\usepackage{amsmath,amsfonts}

\usepackage{array}
\usepackage[caption=false,font=normalsize,labelfont=sf,textfont=sf]{subfig}
\usepackage{textcomp}

\usepackage{url}
\usepackage{wasysym}
\usepackage{verbatim}
\usepackage{booktabs}
\usepackage{graphicx}
\usepackage{color,xcolor}
\usepackage{cite}
\usepackage{bibentry}

\usepackage{natbib}  
\usepackage{caption} 
\usepackage{enumitem}
\usepackage{amssymb} 

\newlist{checklist}{itemize}{1}
\setlist[checklist,1]{
  label={\(\square\)},
  labelwidth=1em,
  labelsep=1em,
  left=0pt,
  itemsep=0pt
}

\usepackage{newfloat}
\usepackage{listings}
\DeclareCaptionStyle{ruled}{labelfont=normalfont,labelsep=colon,strut=off} 
\floatstyle{ruled}
\newfloat{listing}{tb}{lst}{}
\floatname{listing}{Listing}

\title{Graph Mixture of Experts and Memory-augmented Routers for Multivariate Time Series Anomaly Detection}
\author {
    Xiaoyu Huang\textsuperscript{\rm 1,\rm2},
    Weidong Chen\thanks{Corresponding author.}\textsuperscript{\rm 1},
    Bo Hu\textsuperscript{\rm 1}, 
    Zhendong Mao\textsuperscript{\rm 1}
}
\affiliations {

    \textsuperscript{\rm 1}University of Science and Technology of China, Hefei, China\\
    \textsuperscript{\rm 2}Institute of Plasma Physics, Hefei Institutes of Physical Science, Chinese Academy of Sciences, Hefei, China\\
    hhxy@mail.ustc.edu.cn, \{chenweidong, hubo, zdmao\}@ustc.edu.cn
}

\usepackage{bibentry}

\begin{document}

\maketitle

\begin{abstract}

Multivariate time series (MTS) anomaly detection is a critical task that involves identifying abnormal patterns or events in data that consist of multiple interrelated time series. 
In order to better model the complex interdependence between entities and the various inherent characteristics of each entity, the graph neural network (GNN) based methods are widely adopted by existing methods. 
In each layer of GNN, node features aggregate information from their neighboring nodes to update their information. 
In doing so, from shallow layer to deep layer in GNN, original individual node features continue to be weakened and more structural information, \emph{i.e.}, from short-distance neighborhood to long-distance neighborhood, continues to be enhanced. However, research to date has largely ignored the understanding of how hierarchical graph information is represented and their characteristics that can benefit anomaly detection. 
Existing methods simply leverage the output from the last layer of GNN for anomaly estimation while neglecting the essential information contained in the intermediate GNN layers. 
To address such limitations, in this paper, we propose a Graph Mixture of Experts (Graph-MoE) network for multivariate time series anomaly detection, which incorporates the mixture of experts (MoE) module to adaptively represent and integrate hierarchical multi-layer graph information into entity representations. It is worth noting that our Graph-MoE can be integrated into any GNN-based MTS anomaly detection method in a plug-and-play manner. In addition, the memory-augmented routers are proposed in this paper to capture the correlation temporal information in terms of the global historical features of MTS to adaptively weigh the obtained entity representations to achieve successful anomaly estimation. Extensive experiments on five challenging datasets prove the superiority of our approach and each proposed module. 
\begin{links}
  \link{Code}{https://github.com/dearlexie1128/Graph-MoE}
\end{links}

\end{abstract}

\section{Introduction}

Detecting anomalies in multivariate time-series data is crucial for ensuring security and preventing financial loss in industrial applications, where devices like servers and engines are monitored using multivariate time-series entities, organizations need efficient systems to identify potential issues quickly \cite{ren2019time}. However, anomalies are typically rare in both real-world application scenarios and existing public datasets, making data labeling both challenging and costly. Thus, we focus on unsupervised multivariate time series anomaly detection in this paper. 

An effective unsupervised strategy is modeling the dataset into a distribution. As machine learning advances\cite{guo2024benchmarking,gao2023vectorized, gao2020learning, hu2023exploring, xu2024cross, xu2024rule, xu2024multi, ma2024sequential, ma2021poisoning, ma2022tale}, many existing methods \cite{rasul2021multivariate, rasul2021autoregressive} have been explored alongside this strategy to achieve density estimation of the distribution. However, a key challenge with this approach is that many existing methods ignore the interdependencies between time series, which are critical for accurately modeling the data distribution and performing efficient density estimates for anomaly detection. Graph neural network (GNN) based methods have emerged as a promising solution for enhancing anomaly detection in time-series data by modeling the dependencies between entities. \citet{deng2021graph} propose to combine structure learning with GNNs to detect anomalies in MTS. MTAD-GAT \cite{zhao2020multivariate} utilizes graph attention networks to model spatial and temporal dependencies. GReLeN \cite{zhang2022grelen} learns a probabilistic relation graph for multivariate time series and utilizes hidden variables to capture spatial dependencies. \citet{cai2021structural} present a method based on graph neural networks for anomaly detection in dynamic graphs.

\begin{figure}[t]        
\center{\includegraphics[width=0.8\linewidth] {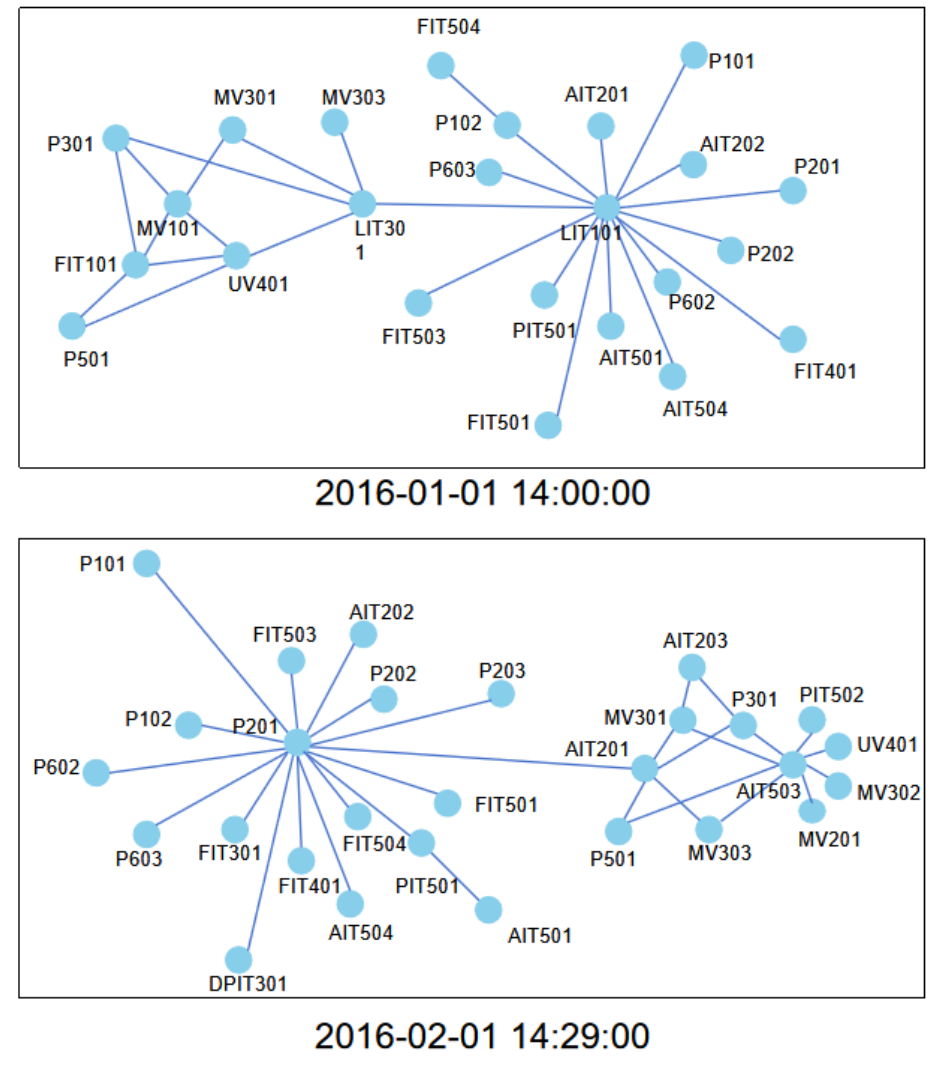}} 
\vspace{-5pt}
\caption{Graph structure of two real-data in MTGFlow \cite{zhou2023detecting}. Each node indicates an entity of multivariate time-series signals of a certain time step.}
\label{fighop}
\end{figure}

Although previous GNN-based methods have achieved promising results, they simply leverage the output from the last layer of GNNs to model the relations between entities. However, as stated in previous analysis works 
\cite{chen2021neural,chen2022heterogeneous,
xu2018representation, he2024ada, wang2024graph, tian2021aspect, yang2024self, yang2024gauss} of GNNs, a common limitation of existing method in MTS anomaly detection is that GNNs are essentially ``homogeneous” across the whole graph, \emph{i.e.}, forcing all nodes to share the same aggregation mechanism, regardless of the differences in their node features or neighborhoods. That might be suboptimal in graph-based relationship modeling, \emph{e.g.}, when some nodes may require information aggregated over longer ranges while others prefer shorter-range local information. As illustrated in the upper example in Fig. \ref{fighop}, the average number of neighborhood nodes within 1-hop, 2-hop, and 3-hop are 2.3, 12.5, and 18.9, respectively, which shows that the network needs to go deep into multi-layer GNNs to establish more global relations. A single-layer GNN can only establish the simplest local relations. Thus, the intermediate layers of GNNs capture information from different multi-hop neighborhoods, \emph{i.e.}, from short-distance to long-distance neighborhoods, which is vital for comprehensively understanding the complex relations and dependencies in MTS data.

In this paper, we propose a Graph Mixture of Experts (Graph-MoE) network, an unsupervised method for multivariate time series (MTS) anomaly detection, to address the aforementioned challenges. Unlike previous GNN-based methods that solely leverage the output from the last layer of GNN to model the interdependences between entities, our Graph-MoE comprehensively utilizes all intermediate information of multi-layer GNNs, that is, it considers the hierarchical node (entity) representations in domains of variable-distance neighborhoods. Specifically, the proposed Graph-MoE incorporates the mixture of experts (MoE) framework to adaptively represent and integrate hierarchical information of different GNN layers into entity representations. In each GNN layer, a specific expert network is implemented with the global-guided entity attention block for the intra-level aggregation of node representations. Moreover, we propose a memory-augmented router, which uses a memory unit to hold the global historical features of the MTS to mine inter-series correlation patterns between time series of the current state, predicting the importance coefficients of experts and facilitating the inter-level aggregation of entity representations. It is worth noting that our Graph-MoE can be integrated into any GNN-based MTS anomaly detection method in a plug-and-play manner. Extensive experiments on five challenging datasets demonstrate the superiority of our approach and each proposed module. In short, our main contributions are summarized as follows:
\begin{itemize}
\item In this paper, we propose a novel Graph Mixture of Experts (Graph-MoE) network for MTS anomaly detection, which extends the expert model at each specific layer of the GNNs. By bringing in experts at different GNN layers with each focusing on different patterns, the Graph-MoE model is able to more fully capture the complex dependencies between entities and the various inherent characteristics of each entity in MTS data.

\item We propose a memory-augmented router module to adaptively integrate the outputs of multi-level GNN layers as well as model the relationship between entities better. Our memory-augmented router module leverages the memory mechanism to store the patterns of global historical temporal features to mine the inter-level time-series correlation, achieving effective inter-level aggregation.

\item Extensive experiments on five public benchmarks (\emph{e.g.}, SWaT, WADI, PSM, MSL, and SMD) demonstrate the effectiveness of our method and each proposed module, \emph{e.g.}, reaching 87.2 and 94.2 on AUROC\%, respectively, on SWaT and WADI datasets.
\end{itemize}

\section{Related Work}

\subsection{Time Series Anomaly Detection}
Time series anomaly detection refers to the identification and discovery of abnormal patterns or behaviors in time series data. However, due to the complexity and volume of time series data, obtaining a well-labeled dataset is challenging \cite{choi2021deep}, unsupervised methods are often used to detect temporal anomalies. 
A typical method is to use auto-encoder (AE) \cite{kingma2013auto}. However, the robustness and generalization capabilities of AEs are limited, which led to the development of VAEs \cite{ su2019robust}. Additionally, there are graph-based methods in unsupervised learning algorithms. MTGNN \cite{wu2020connecting} learns and models relationships between time series variables using a Graph Convolutional Network (GCN). Graph WaveNet \cite{ijcai2019p264} models the spatio-temporal graph structure through graph convolution to detect anomalies. GANF \cite{dai2022graphaugmented} and MTGFlow \cite{zhou2023detecting} leverage graph structures combined with normalizing flow techniques to perform density estimation for anomaly detection in series. 

\begin{figure*}[t]
\centering 
\includegraphics[width=0.92\textwidth]{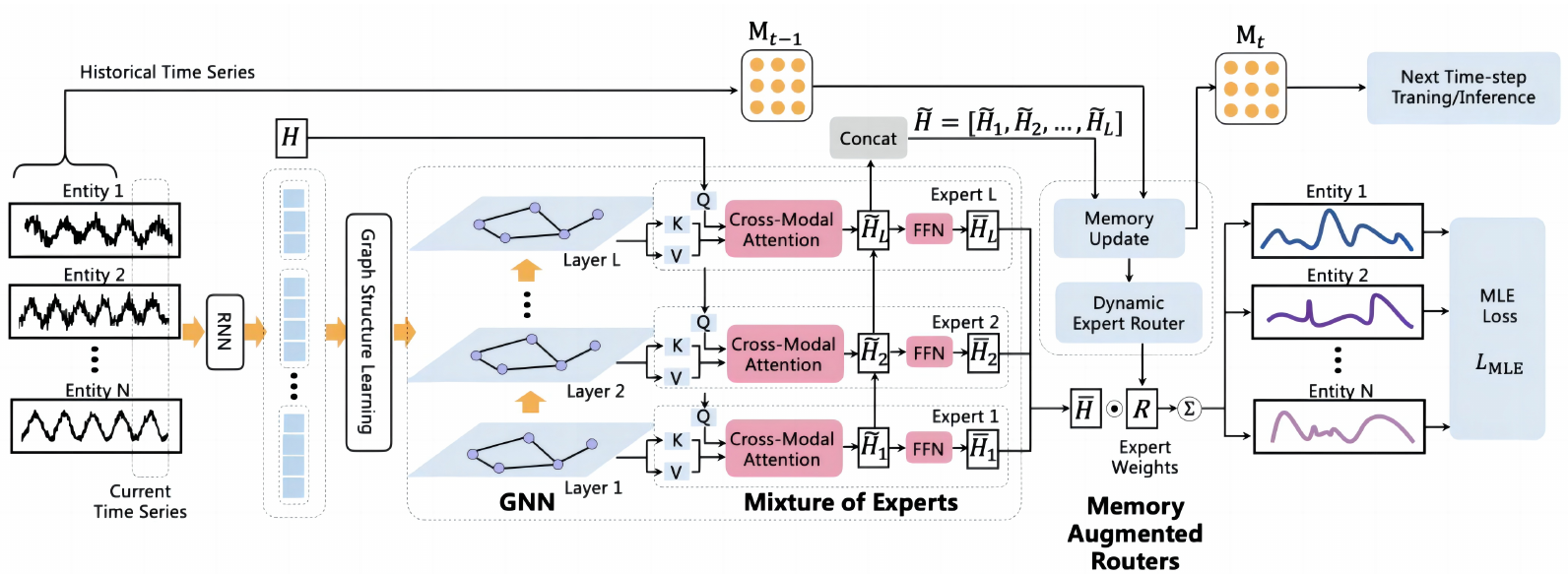} 

\caption{The overview of our proposed Graph-MoE network. It mainly consists of 1) graph mixture of experts and 2) memory-augmented routers.} 
\label{fig1}

\end{figure*}
\subsection{Mixture of Experts}

The Mixture of Experts (MoE) is originally developed as an ensemble method that assigns input data to different ``expert" submodels, each processing a specific part or feature of the input. To adapt MoE for multi-task learning, \citet{ma2018modeling} introduce weight-sharing among multiple tasks. \citet{shazeer2017} introduce the Sparsely-Gated MoE, consisting of thousands of feed-forward sub-networks. Further improvements in the stability and efficiency of sparse MoE models are made by \cite{lewis2021base}. \citet{lepikhin2021gshard} utilize MoE to create V-MoE, a vision transformer with sparse activation. \citet{liu2024multimodal} reveal the Mixture of Multimodal Experts (MoME), enhancing Multimodal Relation Extraction (MRE) through a mixture of hierarchical visual context learners.

\subsection{Memory-augmented Networks}
Growing interest has been attracted to memory-augmented models for improving long-term dependency modeling. LongMem \cite{wang2024augmenting} enhances language models with a memory network to store and retrieve long-term context for better language modeling. In recommendation systems, MA-GNN \cite{ma2020memory} uses a graph neural network for short-term context and a shared memory network for long-term dependencies, while DMAN \cite{tan2021dynamic} segments behavior sequences and uses dynamic memory blocks to combine short-term and long-term user interests for more accurate recommendations. These models highlight the importance of memory mechanisms.

\section{Methodology}
\subsection{Preliminaries}
We present a brief introduction to our baseline in this section to provide a better understanding of our proposed method.

\smallskip

\noindent{\textbf{Normalizing flow.} It is a widely-used method for unsupervised density estimation 
\cite{gudovskiy2022cflow, liu2022unsupervised}, which employs an invertible affine transformation to convert an original distribution into a target distribution. When direct density estimation based on the original data distribution $\mathcal{X}$ is not feasible, alternatives to direct density estimation are to estimate the density of the target distribution $\mathcal{Z}$. We consider the case of a source sample $x \in \mathcal{R}^D$ drawn from the distribution $\mathcal{X}$ and a target distribution sample $z \in \mathcal{R}^D$ drawn from the distribution $\mathcal{Z}$. $z=f_{\theta}(x)$ is a one-to-one mapping between $\mathcal{X}$ and $\mathcal{Z}$. Utilizing the change of variable formula, we can derive that
\begin{equation}
P_{\mathcal{X}}(x) = P_{\mathcal{Z}}(z) \left| {\rm det}\frac{\partial z}{\partial x^\mathrm{T} } \right|.
\end{equation}
Taking advantage of the invertibility of mapping functions and the tractability of Jacobian determinants $\left| {\rm det}\frac{\partial f_{\theta}(x)}{\partial x^\mathrm{T}} \right|$. 
 In flow models, the aim is to achieve $\hat{z} = z$, where $\hat{z} = f_{\theta}(x)$.
When additional conditions $C$ are incorporated, density estimation can be greatly improved. The conditional normalizing flow models represent the mapping as \( z = f_{\theta}(x|C) \). The parameters \( \theta \) of the \( f_{\theta}(x) \) are optimized by maximum likelihood estimation (MLE):
\begin{equation}
\theta^{\ast} = \underset{\theta}{{\arg \max}} ({\rm log}(P_{\mathcal{Z}}(f_{\theta}(x|C))+{\rm log}(\left| {\rm det}\frac{\partial f_{\theta}(x)}{\partial x^\mathrm{T}} \right|)).
\end{equation}}

\noindent\textbf{Notation and Problem Definition.} Multivariate Time Series (MTS) are defined as $x = (x_1, x_2, ..., x_k)$ where $x_i \in \mathcal{R}^L$. For each entity, $k$  signifies the number of entities, while $L$ signifies the number of observations of each entity. We use the z-score to normalize the time series of different entities 
\begin{equation}
\bar{x}_i = \frac{x_i - {\rm mean}(x_i)}{{\rm std}(x_i)},
\end{equation}
where ${\rm mean}(x_i)$ and ${\rm std}(x_i)$ represent the mean and standard deviation of the entity $i$ along the time dimension, respectively. A sliding window with size $T$ and stride size $S$ is used to sample the normalized MTS to preserve the temporal correlations of the original series. The training sample $x^c$ can be obtained by adjusting $T$ and $S$, where $c$ is the number of samples. For simplicity, $x^c$ stands for $x^{cS:cS+T}$.

\subsection{Graph Mixture of Experts}

Our model, depicted in Fig. \ref{fig1}, comprises independent expert networks dedicated to the intra-level aggregation of global-guided entity representations. 

\smallskip

\noindent\textbf{Feature Extraction.} Following previous work \cite{dai2022graphaugmented, zhou2023detecting} for a fair comparison, we model the temporal variations of each entity by using the RNN model, \emph{i.e.,} LSTM \cite{hochreiter1997long} and GRU \cite{cho-etal-2014-learning}. 
\begin{equation}
H_k^t = {\rm RNN}(x_k^t, H_k^{t-1}),
\end{equation} 
where $H_k^t$ is the hidden RNN features for a window sequence of entity $k$, $x_c$ at time $t \in [cS:cS+T)$. The combination of entities $H_k^t$ for $\forall k$ forms the initial input $H$ of our proposed Graph-MoE network.

\smallskip

\noindent\textbf{Graph Construction.} Due to the mutual and evolving dependence among entities, we utilize self-attention to learn a dynamic graph structure by treating entities at a certain time window of multivariate time series as graph nodes.  
For each input window sequence $x^c$, two linear transformations are performed, and then, the pairwise relationship $e_{ij}^c$ at time window $c$ between nodes $i$ and $j$ is determined as:
\begin{equation}
e^c_{ij} = \big(\phi_e^1(x^c_i)\big)\cdot \big(\phi_e^2(x^c_j)\big)^\top,
\end{equation}
where $\phi_e^1$ and $\phi_e^2$ are two linear transformations. To quantify the relationship between node $i$ and $j$, the attention score $a_{ij}^c$ of the matrix $A^c$ is calculated by 
\begin{equation}
a_{ij}^c = \frac{\exp(e_{ij}^c)}{\sum_{j=1}^{K} \exp(e_{ij}^c)}.
\end{equation}
The attention matrix $A^c$ inherently represents the mutual dependencies among the entities, which is treated as the adjacency matrix of the constructed graph, consequently. With time series inputs continually changing, $A^c$ also evolves to capture the dynamic interdependence between them.

\smallskip

\noindent\textbf{Graph Updating.} 
To derive the temporal embedding $H_t^l$ for the $l$-th GNN layer of all entities at $t$ with incorporating the temporal features $H_t^{l-1}$ of the last graph layer, a graph convolution operation is performed through the learned graph $A^c$. 
Inspired by previous work GANF \cite{dai2022graphaugmented} and MTGFlow \cite{zhou2023detecting}, we also find that the history information of the node itself helps enhance temporal relationships of time series. Hence, the $l$-th GNN layer of spatio-temporal condition at time $t$ is defined as:
\begin{equation}
H^{l}_t = {\rm ReLU}(A^c H^{l-1}_t W_1+H^{l-1}_{t-1}W_2)\cdot W_3,
\end{equation}
where $W_1$ and $W_2$ are graph convolution and history information weights, respectively. $W_3$ is used to improve the expression ability of condition representations. Superscript $l$ indicates the output of the $l$-th GNN layer.  
The spatio-temporal condition $H^{l}_c$ for window $c$ of the $l$-th GNN layer is the concatenation of $H^{l}_t$ along the time axis.

\smallskip

\noindent\textbf{Mixture of Experts.} 
A mixture of experts (MoE) is proposed and introduced to explore whether distinct global temporal features may favor graph information encoded at different levels. 
A mixture of experts dynamically combines the outputs of sub-models known as ``experts" via a weighting function known as the ``router". The proposed mixture of experts is formulated as follows:
\begin{equation}
C = \sum_{l=1}^{L} R_l \cdot f_l(H^l) = \sum_{l=1}^{L} R_l \cdot \bar{H}^l, \label{eq8}
\end{equation}
where $f_l(\cdot)$ denotes the $l$-th expert function, $\bar{H}^l$ is the expert-aligned entity features of the $l$-th graph layer, 
and $R=[R_1;R_2; ...;R_L]$ is the routing weights for the $L$ experts. For each entity $x_i$, since each GNN layer gathers information from directly connected entities, stacking multiple GNN layers enables the learning of neighborhood relationships between entities over long distances. Thus, each layer of our Graph-MoE has a specific expert network that independently captures features at different levels that focus on the different range of neighborhoods. 

Our MoE takes the current temporal embedding $H$ obtained from the RNN encoder as queries and the spatio-temporal condition $H^l$ extracted from the $l$-th GNN layer as key-value pairs to aggregate node features within each layer. Formally, the cross-attention block operates as follows:

\begin{equation}
\begin{aligned}
\tilde{H}^l &= \text{LN}(\text{Attention}(H, H^l)) |_{Q:H,\{K,V\}:H^l},
\end{aligned}
\end{equation} 
where ${\rm LN}$ is the layer normalization operation. To enhance the capabilities of intra-layer feature representations, we employ the FFN and the ReLU activation function in the expert network. The expert-aligned entity features $\bar{H}^l$ are:
\begin{equation}
\begin{aligned}
\bar{H}^l &= \text{FFN}_l(\tilde{H}^l) =\phi^2_l \Big({\rm ReLU}\big(\phi^1_l (\tilde{H}^l)\big)\Big),
\end{aligned}
\end{equation}where $\phi_l^1$ and $\phi_l^2$ refer to the multi-layer perceptron (MLP). Then we respectively concatenate the outputs obtained from the attention mechanism, forming hierarchical global-guided temporal features $\tilde{\mathbf{H}}$: 
\begin{equation}
\tilde{\mathbf{H}} = [\tilde{{H}}^1, \tilde{{H}}^2, \ldots, \tilde{{H}}^L].
\end{equation}

For current time step $t$, $\tilde{\mathbf{H}}$ is specified as $\tilde{\mathbf{H}}_t$, which is then used for memory updates and dynamic routers generation.

\subsection{Memory-augmented Routers}

The memory-augmented router is designed to extract features from the current temporal slice while also capturing the characteristics of the entire historical data. These extracted features are then utilized by the module to predict the importance coefficients of experts of different time series. 

To exploit the characteristics of memory, inspired by \cite{liu2022memory, chen2020generating}, we propose memory-augmented routers to enhance the weighted combination of GNNs' multi-layer outputs. In doing so, the memory module uses a matrix to store the global historical temporal features. In the process of updating, the matrix is revised step-by-step incorporating the output from previous steps. Then, at time step $t$, the matrix from the previous step, $M_{t-1}$, is functionalized as the query, and its concatenations with the previous output serve as the key and value to feed the multi-head attention module.  
The entire process is formulated as follows:
\begin{equation}
\begin{aligned}
Y &= [M_{t-1}; \tilde{\mathbf{H}}_t],\\
Z &= \text{Attention}(M_{t-1}, Y) |_{Q:M_{t-1},\{K,V\}:Y},
\end{aligned}
\end{equation}
where $[M_{t-1}; \tilde{\mathbf{H}}_t]$ is the row-wise concatenation of $M_{t-1}$ and $\tilde{\mathbf{H}}_t$. 
Considering that the memory module is performed in a recurrent manner along with the estimation process, it potentially suffers from gradient vanishing and exploding. We therefore introduce residual connections and a gate mechanism. The former is formulated as
\begin{figure}
    \centering
    \includegraphics[width=0.6\linewidth]{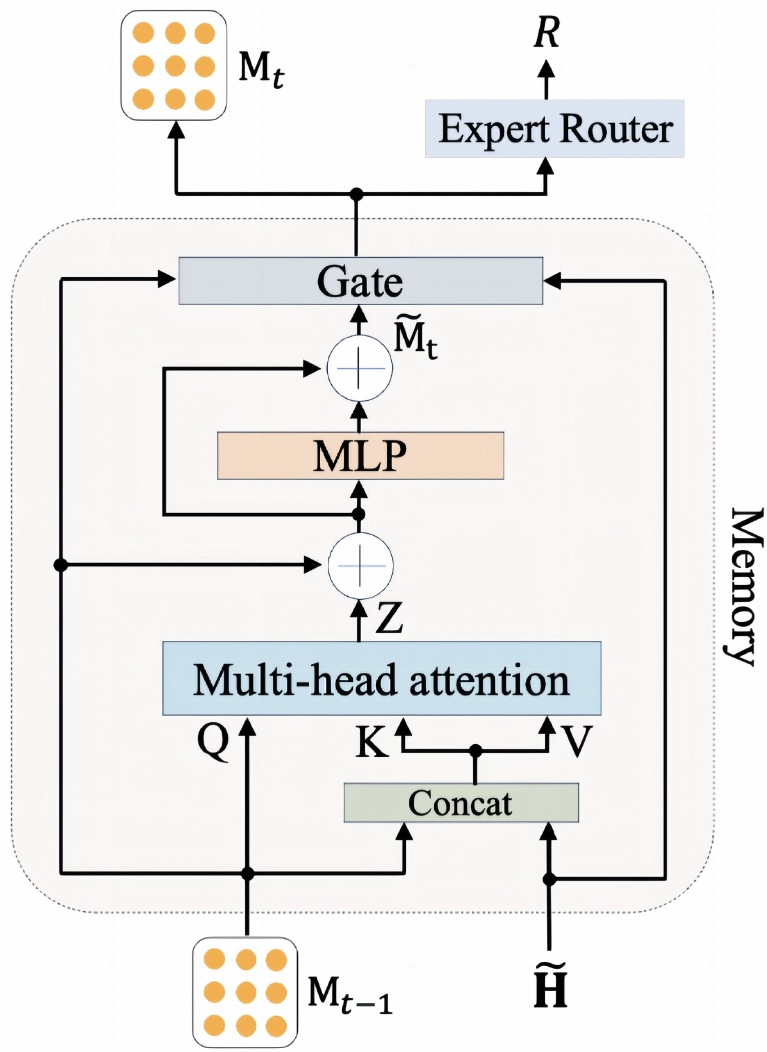}
    \caption{The framework of our memory-augmented routers.}
    \label{fig3_3}
\end{figure}
\begin{equation}
\tilde{M}_t = \phi_M(Z + M_{t-1}) + Z + M_{t-1},
\end{equation}
where $\phi_M$ refers to the multi-layer perceptron (MLP). The detailed structure of the gate mechanism in the memory module is shown in Fig. \ref{fig3_3}, where the forget and input gates are applied to balance the inputs from $M_{t-1}$ and $\tilde{\mathbf{H}}_t$, respectively. The forget and input gates are formalized as
\begin{equation}
G^f_t = \tilde{\mathbf{H}}_{t}W^f + \tanh(M_{t-1}) \cdot U^f,
\end{equation}
\begin{equation}
G^i_t = \tilde{\mathbf{H}}_{t}W^i + \tanh(M_{t-1}) \cdot U^i,
\end{equation}
where $W^f$ and $W^i$ are trainable weights for $\tilde{\mathbf{H}}_{t}$ in each gate; and similarly, $U^f$ and $U^i$ are the trainable weights for $M_{t-1}$ in each gate. The final output of the gate mechanism is formalized as follows:
\begin{equation}
M_t = \sigma(G^f_t) \odot M_{t-1} + \sigma(G^i_t) \odot \tanh(\tilde{M}_t),
\end{equation}
where $\odot$ refers to the Hadamard product, $\sigma$ is the sigmoid activation, and $M_t$ is the output of the memory module at step $t$, which contains both the global historical temporal information and the current temporal features. We then update the memory by replacing $M_{t-1}$ with ${M}_t$ in the next step.

The output $\phi_R(M_t)$, derived from the memory module, is then fed into the expert router, which is a linear transformation $\phi_R$ of the input followed by a softmax layer:  
\begin{align}
R = {\rm Softmax}{(\phi_R(M_t))},
\end{align}
where the output $R$ preserves the predicted expert weights through the memory-augmented router. Finally, we leverage Eq. (\ref{eq8}) to generate the spatio-temporal condition $C$.

\section{Experiment}
\subsection{Experiment Setup}
\noindent\textbf{Datasets.} Following GANF \cite{dai2022graphaugmented} and MTGFlow \cite{zhou2023detecting}, we conduct experiments on five widely-used public datasets for MTS anomaly detection. The details are illustrated as follows: SWaT \cite{mathur2016swat} consists of 51 sensors from an industrial water treatment plant, including 41 attacks over four days. WADI \cite{ahmed2017wadi} is a collection of 123 sensor and actuator data from the WADI testbed, at the frequency of one second. PSM \cite{abdulaal2021practical} is a collection of multiple application server nodes at eBay with 25 features, providing ground truths created by experts over 8 weeks. MSL \cite{hundman2018detecting} is a collection of the sensor and actuator data of the Mars rover with 55 dimensions. SMD \cite{su2019robust} is a 5-week-long dataset that is collected from a large Internet company with 38 dimensions.

\smallskip

\noindent\textbf{Evaluation Metrics.} Following previous studies \cite{dai2022graphaugmented, zhou2023detecting}, Graph-MoE is designed to detect anomalies at the window level, where a window is labeled as abnormal if any time point within it is identified as anomalous. The performance of the model is evaluated using the Area Under the Receiver Operating Characteristic Curve (AUROC).

\smallskip

\noindent\textbf{Implementation Details.} We set the window size as 60 and the stride size as 10. During the training stage, we utilize an Adam optimizer. For the SWaT dataset, we use a single flow block with a batch size of 640 and a learning rate of 0.005. The batch size of the WADI dataset is 320, and the learning rate is 0.0035. Other datasets employ two flow blocks with a batch size of 256 and a learning rate of 0.002. We follow the dataset setting of previous work \cite{zhou2023detecting, dai2022graphaugmented} and split the original testing dataset by 60\% for training, 20\% for validation, and 20\% for testing in SWaT. For other datasets, the training split contains 60\% data, and the test split contains 40\% data. All our experiments are conducted on GPU, running for 80 epochs.

\smallskip

\noindent\textbf{Baselines.} To demonstrate the superiority of our method, we compare it with current competitive approaches, including the state-of-the-art methods that encompass both semi-supervised and unsupervised approaches to anomaly detection. Notably, semi-supervised methods include DeepSAD \cite{Ruff2020Deep} and DROCC \cite{goyal2020drocc}. Besides, unsupervised methods include DeepSVDD \cite{ruff2018deep}, ALOCC \cite{sabokrou2020deep}, USAD \cite{audibert2020usad}, DAGMM \cite{zong2018deep}, GANF \cite{dai2022graphaugmented} and MTGFlow \cite{zhou2023detecting}.


\begin{table*}[t]
\centering
\caption{Comparison with the state-of-the-art methods in anomaly detection on five challenging datasets, \emph{i.e.},
SWaT, WADI, PSM, MSL, and SMD. The best results are highlighted in bold.}
\label{table1}
\resizebox{0.85\linewidth}{!}{
\begin{tabular}{ccccccc}
\toprule
\multirow{2}*{Method} &\multirow{2}*{Venue} &\multicolumn{5}{c}{Datasets} \\
\cmidrule{3-7}
& & {SWaT} & {WADI} & {PSM} & {MSL} & {SMD} \\
\midrule
DeepSVDD \cite{ruff2018deep} & ICML2018 & 66.8$\pm$2.0 & 83.5$\pm$1.6 & 67.5$\pm$1.4 & 60.8$\pm$0.4 & 75.5$\pm$15.5 \\
DAGMM \cite{zong2018deep} & ICLR 2018 & 72.8$\pm$3.0 & 77.2$\pm$0.9 & 64.6$\pm$2.6 & 56.5$\pm$2.6 & 78.0$\pm$9.2 \\
ALOCC \cite{sabokrou2020deep} & TNNLS 2020 & 77.1$\pm$2.3 & 83.3$\pm$1.8 & 71.8$\pm$1.3 & 60.3$\pm$0.9 & 80.5$\pm$11.1 \\
DROCC \cite{goyal2020drocc} & ICML 2020 & 72.6$\pm$3.8 & 75.6$\pm$1.6 & 74.3$\pm$2.0 & 53.4$\pm$1.6 & 76.7$\pm$8.7 \\
DeepSAD \cite{Ruff2020Deep} & ICLR 2020 & 75.4$\pm$2.4 & 85.4$\pm$2.7 & 73.2$\pm$3.3 & 61.6$\pm$0.6 & 85.9$\pm$11.1 \\
USAD \cite{audibert2020usad} & KDD 2020 & 78.8$\pm$1.0 & 86.1$\pm$0.9 & 78.0$\pm$0.2 & 57.0$\pm$0.1 & 86.9$\pm$11.7 \\
GANF \cite{dai2022graphaugmented} & ICLR 2022 & 79.8$\pm$0.7 & 90.3$\pm$1.0 & 81.8$\pm$1.5 & 64.5$\pm$1.9 & 89.2$\pm$7.8 \\
MTGFlow \cite{zhou2023detecting} & AAAI 2023 & 84.8$\pm$1.5 & 91.9$\pm$1.1 & 85.7$\pm$1.5 & 67.2$\pm$1.7 & 91.3$\pm$7.6 \\
\midrule
\rowcolor{gray!15}%
Ours (Graph-MoE) &  & \textbf{87.2$\pm$1.3} & \textbf{94.2$\pm$0.8} & \textbf{88.0$\pm$0.7} & \textbf{72.1$\pm$1.1} & \textbf{93.3$\pm$5.6} \\
\bottomrule
\end{tabular}}
\end{table*}

\subsection{Main Comparison}
As shown in Table \ref{table1}, we compare Graph-MoE with the aforementioned baselines and report the average performance and standard deviations on AUROC scores through five random repetitions.  
The results show that our Graph-MoE reaches an exceptionally high AUROC score compared to the other eight baselines. 
Note that the standard deviation observed in the SMD arises from the fact that it encompasses 28 distinct sub-datasets. For each sub-dataset, we evaluate the performance and subsequently calculate the average of all these individual results. From the results of Table \ref{table1}, we have the following observations.

\smallskip

\noindent$\bullet$ Compared with semi-supervised methods, our method achieves significant improvements on metric AUROC\%. For instance, on the SWaT dataset, our method achieves improvements over DeepSAD and DROCC, +15.6\% and +20.1\%, respectively. The reason for such remarkable improvements is that, unlike our Graph-MoE, DROCC projects all training samples into a hypersphere, hindering their ability to accurately distinguish between normal and abnormal samples. DeepSAD relies on the distance to the center to detect anomalies, and its effectiveness is limited when handling complex anomaly patterns.

\smallskip

\noindent$\bullet$ Our model also achieves better performance than the unsupervised methods in the baseline. For example, on the WADI dataset, our model demonstrate a +22.0\% increase compared to DAGMM, +4.3\% compared to GANF, and +2.5\% compared to MTGFlow. The reason is that our Graph-MoE leverages the multi-layer GNN to comprehensively capture information from short-distance to long-distance neighborhoods, enhancing the understanding of complex relationships in multivariate time series. By introducing the Mixture of Experts (MoE) and the Memory-augmented Routers (MAR), the model adaptively selects suitable experts based on spatio-temporal features, effectively combining global and local information for precise detection.

\begin{table}[t]
\centering
\caption{\label{table2}The results of ablation studies on the SWaT dataset to discuss the number of experts, which is
the important hyperparameter of our method.}
\resizebox{0.62\linewidth}{!}{
\begin{tabular}{ccc}
\toprule
\multirow{2}*{\# of experts}   &\multicolumn{2}{c}{Datasets} \\
\cmidrule{2-3}
&{SWaT} &{WADI} \\        
\midrule
1 & 86.2$\pm$0.7 & 92.5$\pm$1.2 \\
2 & 86.8$\pm$0.4 & 93.4$\pm$0.6 \\
\rowcolor{gray!15}
3 & \textbf{87.2$\pm$1.3} & \textbf{94.2$\pm$0.8} \\
4 & 85.6$\pm$2.1 & 92.6$\pm$0.7 \\
\bottomrule
\end{tabular}}
\vspace{-5pt}
\end{table}
\subsection{Ablation Studies}

\noindent\textbf{Effectiveness on the number of Graph-MoE layers.} 
In our proposed method Graph-MoE, the most critical parameter is the number of layers. We explore the influence of different layers of Graph-MoE on anomaly detection performance, and the results are shown in Table \ref{table2}. We can observe that the AUROC fraction increases gradually with the increase in the number of layers and reaches a peak at the third-layer Graph-MoE. However, as the number of layers increases to 4, the performance declines. 

The reason is that when the number of layers is set to 1 and 2, the model can only aggregate a restricted amount of neighborhood information, which is insufficient for capturing complex long-distance relationships. 
Moreover, an excessive number of layers can lead to the occurrence of over-smoothing, which reduces the distinctiveness between node features, diminishes diversity and ultimately weakens the model's representation ability. The results show that a reasonable selection of layers can significantly improve the detection ability of the model.
 
\smallskip

\begin{table}
\centering
\caption{\label{table3} The results of ablation studies on SWaT and WADI datasets to discuss the effectiveness of our proposed components. MoE and MAR stand for the mixture of experts and memory-augmented routers, respectively.}
\resizebox{0.62\linewidth}{!}{
\begin{tabular}{cccc}
\toprule
\multicolumn{2}{c}{Components} &\multirow{2}*{SWaT} &\multirow{2}*{WADI}  \\     
\cmidrule{1-2}
MoE &MAR &  & \\
\midrule
{$\times$}&{$\times$} &85.5$\pm$1.1&92.2$\pm$0.2\\
{$\times$}&{$\checkmark$} & 86.2$\pm$1.2&92.9$\pm$1.0\\
{$\checkmark$}&{$\times$} & 86.8$\pm$1.2&93.5$\pm$0.4\\
\rowcolor{gray!15}
{$\checkmark$}&{$\checkmark$} &  \textbf{87.2$\pm$1.3} & \textbf{94.2$\pm$0.8} \\
\bottomrule
\end{tabular}}
\vspace{-5pt}
\end{table}

\begin{figure*}[t]
    \centering
    \includegraphics[width=1.0\linewidth]{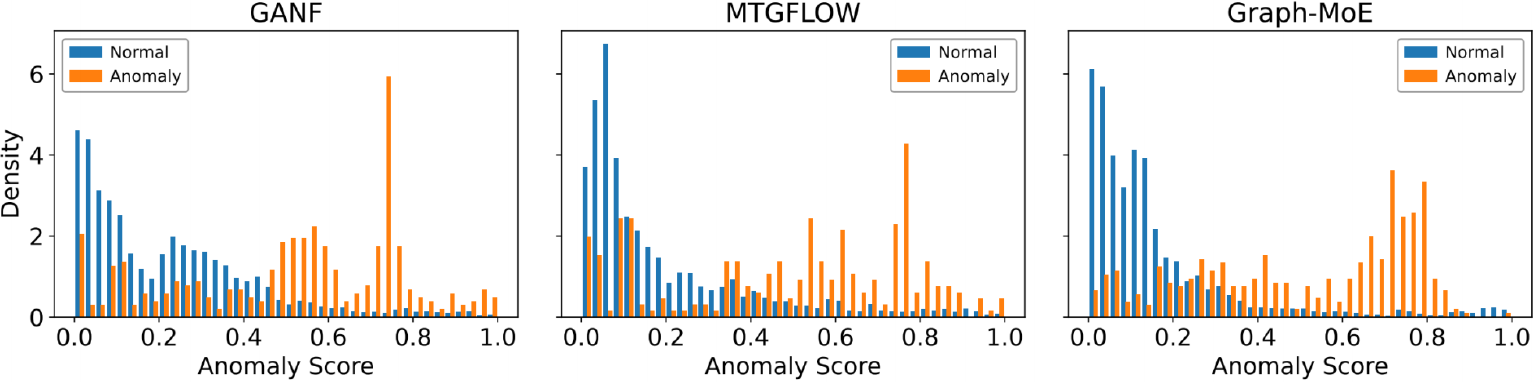}
    \caption{Distribution of anomaly scores for normal and anomaly data across GANF, MTGFlow, and Graph-MoE models.}
     \label{fig3}
\end{figure*}

\noindent\textbf{Discussion on each proposed module.} To further investigate the contribution of each component in our method, we conduct a set of ablation studies. The results are shown in Table \ref{table3}. In this table, ``$\times$" indicates that the module is not used, while ``$\checkmark$" indicates that the module is utilized. The components MoE and MAR represent the mixture of experts and memory-augmented routers, respectively. 
Initially, without the inclusion of MAR and MoE, there are modest performance improvements of 0.8\% and 0.3\% on SWaT and WADI over MTGFlow, respectively, which suggests that the base graph structure alone is somewhat effective in capturing spatio-temporal information. However, when MAR and MoE are introduced separately, both contribute to further performance gains, with MoE having a more pronounced impact. The results could be attributed to MoE’s ability to better integrate expert knowledge from different GNN layers, enhancing the model's representational capacity. When MAR and MoE are both introduced into our model, their synergistic effect leads to substantial performance improvements of 2.8\% on the SWaT dataset and 2.5\% on the WADI dataset, which underscores the effectiveness of combining MAR's global historical feature extraction with MoE's integration of multi-level information in addressing complex time series anomaly detection challenges.

\smallskip

\noindent\textbf{Discussion on the plug-and-play properties.} To evaluate the plug-and-play capability of our model, we select three representative graph-based anomaly detection methods, namely GANF \cite{dai2022graphaugmented}, MTGFlow \cite{zhou2023detecting}, and USD \cite{liu2024usd}. As shown in Table \ref{table4}, we apply Graph-MoE upon these three baselines. Specifically, after incorporating Graph-MoE, GANF's performance increases from 79.8 to 82.6, representing a gain of 3.5 percentage points, the largest improvement observed. MTGFlow's performance rises from 84.8 to 87.2, with a 2.8\% increase. Similarly, USD's performance improves from 90.2 to 92.3, an increase of 2.3 percentage points. This consistent performance enhancement proves that Graph-MoE effectively strengthens anomaly detection capabilities across different graph-based models, demonstrating the strong plug-and-play capabilities of our proposed Graph-MoE.

\begin{table}
\centering
\caption{\label{table4} We integrate our Graph-MoE into the three baseline methods of GANF, MTGFlow, and USD, and the results show the superiority of our model.}
\resizebox{0.4\textwidth}{!}{
\begin{tabular}{l|c|c}
\toprule
\textbf{Methods} & \textbf{SWaT} & \textbf{Improvement($\Delta$\%)}\\
\midrule 
GANF  & 79.8$\pm$0.7  \\
\rowcolor{gray!15} %
GANF+Graph-MoE &\textbf{82.6$\pm$0.7} &$\Delta$\%=3.5 \\
\midrule
MTGFlow & 84.8$\pm$1.5  \\
\rowcolor{gray!15} %
MTGFlow+Graph-MoE &\textbf{87.2$\pm$1.3} &$\Delta$\%=2.8 \\
\midrule
USD & 90.2$\pm$0.9 \\
\rowcolor{gray!15} %
USD+Graph-MoE &\textbf{92.3$\pm$1.4} &$\Delta$\%=2.3 \\
\bottomrule
\end{tabular}}
\end{table}
\subsection{Qualitative Results}
\begin{figure}
    \centering
    \includegraphics[width=0.9\linewidth]{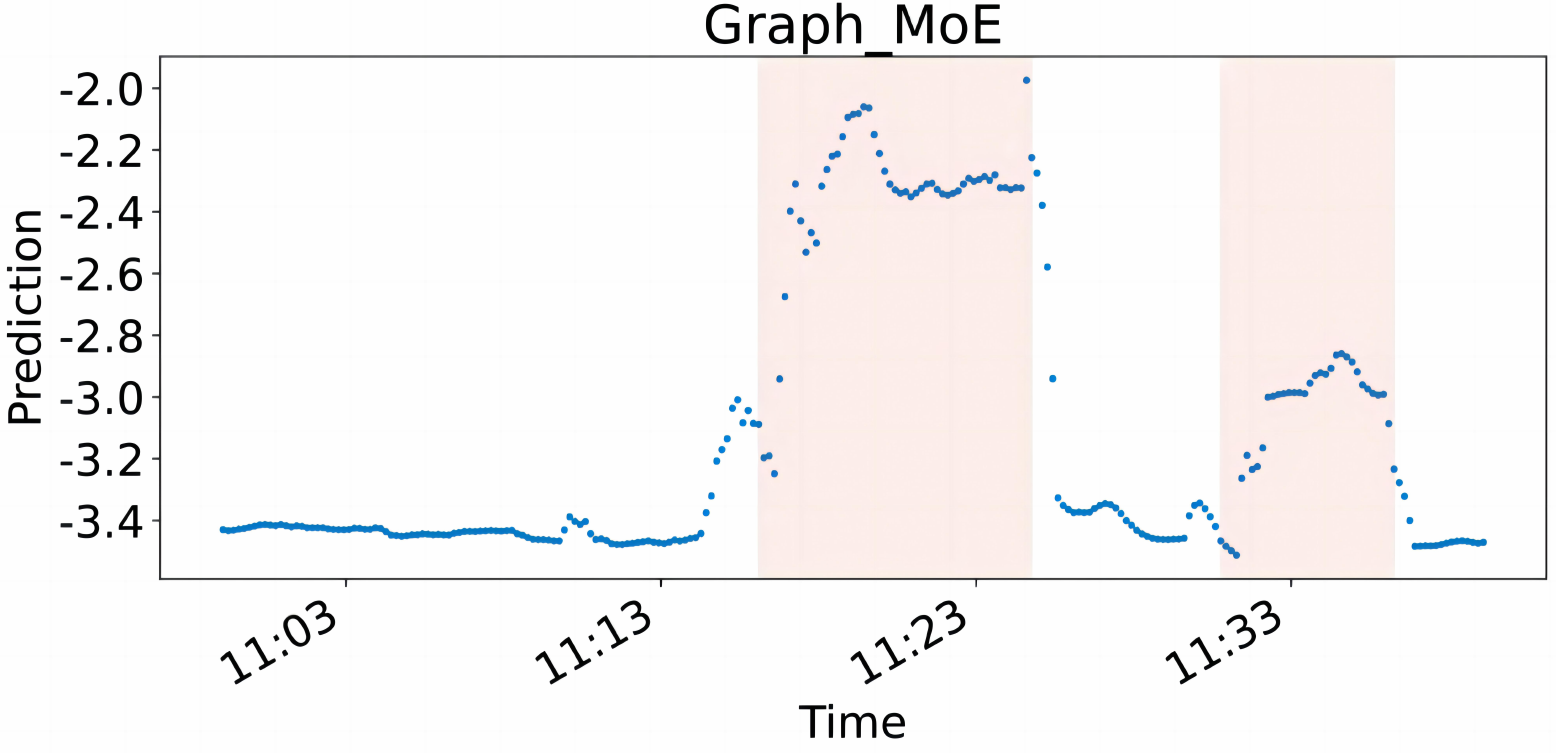}
    \caption{Anomaly detection over time by our Graph-MoE.}
    \label{fig2}
    \vspace{-5pt}
\end{figure}

Moreover, we visualize the distributions of data on the SWAT dataset to examine the effectiveness of our method. To investigate the anomaly discrimination ability of Graph-MoE, we present the normalized anomaly score for GANF, MTGFlow and Graph-MoE in Fig. \ref{fig3}. As it is displayed, for normal series, anomaly scores of Graph-MoE are more centered at 0 than those of GANF and MTGFlow, and the overlap areas of normal and abnormal scores are also smaller in Graph-MoE, reducing the false positive ratio. This larger score discrepancy corroborates that Graph-MoE has superior detection performance. Fig. \ref{fig2} shows the point plot of Graph-MoE anomalies predicted outputs over time. The X-axis displays the progression of time, while the Y-axis indicates the log-likelihood of the anomaly. Abnormal series are highlighted with a red background. This fluctuation in log likelihoods confirms that our proposed Graph-MoE has capabilities for detecting anomalies by identifying low-density regions within the modeled distribution. 

\section{Conclusion}
In this work, we proposed Graph-MoE, an unsupervised-learning based method for anomaly detection in multivariate time series data. Unlike previous methods, we employ a multi-layer, multi-expert learning architecture to dynamically model the  complex interdependencies between entities and the various inherent characteristics of each entity at different levels, capturing both local and global information and enhancing feature representations of each entity. We also propose Memory-augmented Routers, which leverage the memory mechanism to store the patterns of global historical temporal features to predict the importance coefficients of
experts, therefore facilitating effective inter-level aggregation. Experimental results demonstrate that the proposed framework achieves state-of-the-art performance.

\section{Acknowledgements}
This work is supported by the National Key Research and Development Program under Grant No. 2023YFC3303800, and the National Science Fund for Excellent Young Scholars under Grant 62222212.

\bibliography{aaai25}
\clearpage

\end{document}